\documentclass[a4paper,fleqn]{cas-dc}

\usepackage[numbers]{natbib}
\usepackage{amssymb, pifont}
\newcommand{\Xmark}{\ding{55}}
\newcommand{\chmark}{\ding{51}}
\usepackage{paralist}
\usepackage{booktabs}
\usepackage{caption}
\usepackage{array}
\usepackage{threeparttable}
\usepackage{graphicx}
\usepackage{multirow}

\def\tsc#1{\csdef{#1}{\textsc{\lowercase{#1}}\xspace}}
\tsc{WGM}
\tsc{QE}
\begin{document}
\let\WriteBookmarks\relax
\def\floatpagepagefraction{1}
\def\textpagefraction{.001}
\let\printorcid\relax
\shorttitle{}    

\shortauthors{}  

\title [mode = title]{Generating Vision-Language Navigation Instructions Incorporated Fine-Grained Alignment Annotations}

\author[1,2]{Yibo Cui}%[<options>]

\ead{yb.cui@outlook.com}

\affiliation[1]{organization={Defense Innovation Institute, Chinese Academy of Military Science},
            city={Beijing},
%          citysep={}, % Uncomment if no comma needed between city and postcode
            postcode={100071}, 
            country={China}}
\affiliation[2]{organization={Tianjin Artificial Intelligence Innovation Center},
            city={Tianjin},
%          citysep={}, % Uncomment if no comma needed between city and postcode
            postcode={300450}, 
            country={China}}

\author[1,2]{Liang Xie}%[]
            
\author[2]{Yu Zhao}%[<options>]

\author[2,3]{Jiawei Sun}%[<options>]

\affiliation[3]{organization={School of Mechanical Engineering, Tianjin University},
            city={Tianjin},
%          citysep={}, % Uncomment if no comma needed between city and postcode
            postcode={300354}, 
            country={China}}

\author[1,2]{Erwei Yin}%[<options>]

\cormark[1]

\cortext[1]{Corresponding author}

\ead{yinerwei1985@gmail.com}


\begin{abstract}
Vision-Language Navigation (VLN) enables intelligent agents to navigate environments by integrating visual perception and natural language instructions, yet faces significant challenges due to the scarcity of fine-grained cross-modal alignment annotations.
Existing datasets primarily focus on global instruction-trajectory matching, neglecting sub-instruction-level and entity-level alignments critical for accurate navigation action decision-making. To address this limitation, we propose FCA-NIG, a generative framework that automatically constructs navigation instructions with dual-level fine-grained cross-modal annotations. 
In this framework, an augmented trajectory is first divided into sub-trajectories, which are then processed through GLIP-based landmark detection, crafted instruction construction, OFA-Speaker based R2R-like instruction generation, and CLIP-powered entity selection, generating sub-instruction-trajectory pairs with entity-landmark annotations.
Finally, these sub-pairs are aggregated to form a complete instruction-trajectory pair.
The framework generates the FCA-R2R dataset, the first large-scale augmentation dataset featuring precise sub-instruction-sub-trajectory and entity-landmark alignments.
Extensive experiments demonstrate that training with FCA-R2R significantly improves the performance of multiple state-of-the-art VLN agents, including SF, EnvDrop, RecBERT, and HAMT. Incorporating sub-instruction-trajectory alignment enhances agents’ state awareness and decision accuracy, while entity-landmark alignment further boosts navigation performance and generalization. 
These results highlight the effectiveness of FCA-NIG in generating high-quality, scalable training data without manual annotation, advancing fine-grained cross-modal learning in complex navigation tasks. 
\end{abstract}

%\nocite{*}

% Keywords

\maketitle

% Main text

\section{Introduction}
Vision-Language Navigation (VLN) represents a pivotal task within embodied artificial intelligence, encapsulating the integration of perceptual understanding from visual input and semantic interpretation derived from natural language instructions to enable intelligent agents to navigate physical or virtual environments~\cite{DBLP:conf/cvpr/AndersonWTB0S0G18, ALFRED}.
VLN has attracted burgeoning research interest from communities of computer vision, natural language processing, and robotics due to its broad application prospects~\cite{DBLP:conf/acl/GuSWTW22, DBLP:journals/pami/DeSouzaK02, DBLP:journals/jair/FrancisKLLNO22}.
In VLN tasks, cross-modal alignment is one central challenge for decision-making since the agent requires dynamically ground the received instructions to the surrounding visual observations. 
For instance, given the instruction "Walk beside the outside doors and behind the chairs across the room. Turn right and walk up the stairs. Stop on the seventh step.", the agent needs to find "outside doors", "chairs", "stairs" etc. (entity level) in nearby environments to help make action decisions and disregard the initial sub-instructions "Walk beside the outside doors" (sub-instruction level) after moving several steps. 
However, due to the scarcity of VLN datasets with these fine-grained alignment signals, previous methods have focused on global matching between instructions and complete visual trajectories (instruction level), leaving cross-modal alignment a significant challenge. 

\begin{figure}
    \centering
    \includegraphics[width=0.5\textwidth]{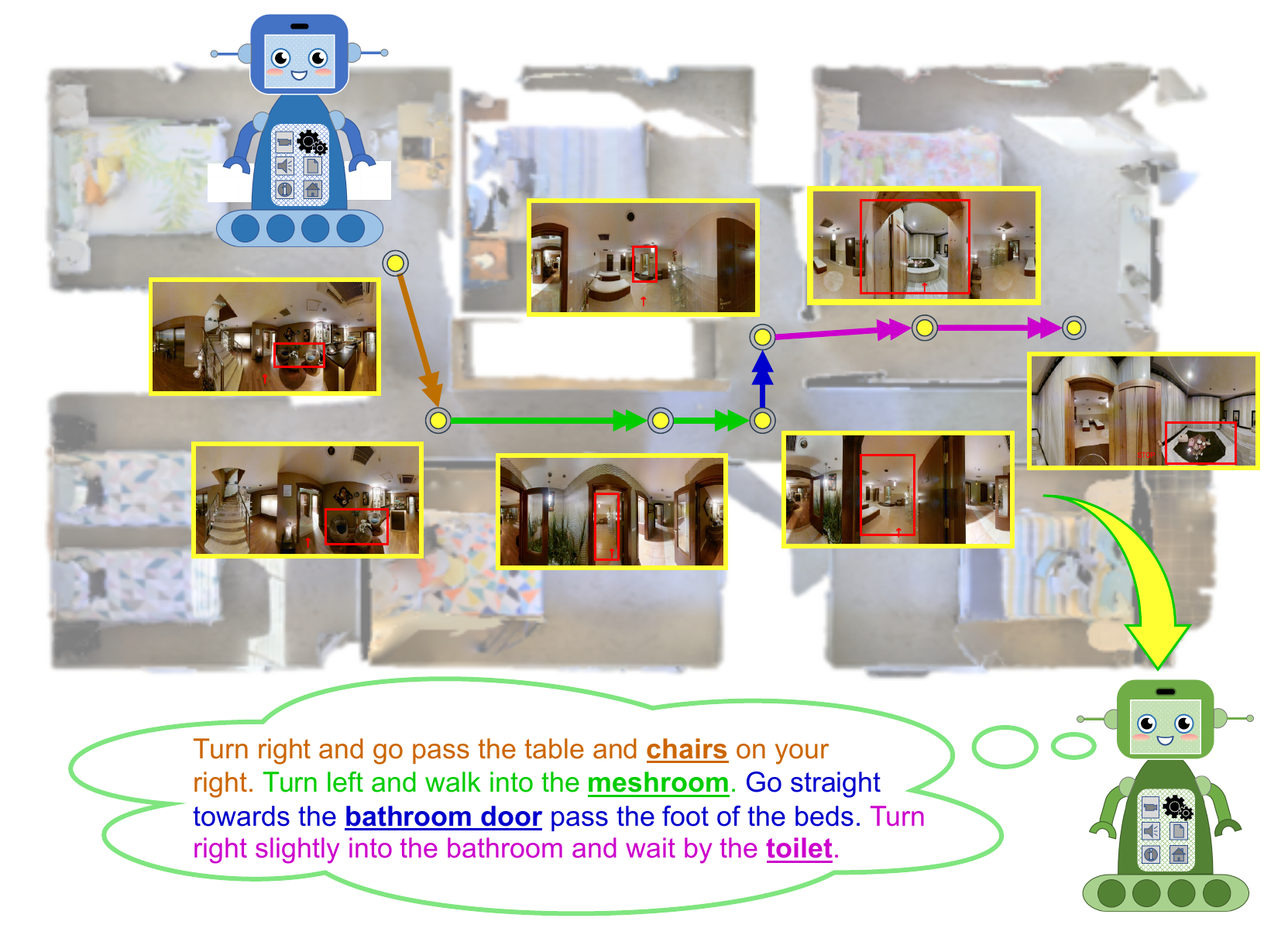}
    \caption{An illustration of our speaker-follower framework. The follower (blue) navigates through a 3D photorealistic environment, while the speaker (green) generates grounded descriptions of navigation trajectories along with fine-grained alignment annotations.}
    \label{fig:example}
\end{figure}

Several prior works have explored fine-grained cross-modal alignment in VLN. Self-Monitoring~\cite{DBLP:conf/iclr/MaLWAKSX19} located completed or ongoing sub-instructions through visual-textual co-grounding and a progress monitor module during navigation. AuxRN~\cite{DBLP:conf/cvpr/Zhu0CL20} reported the percentage of completed procedures at each step using an auxiliary reasoning objective for progress estimation. Qi et al.~\cite{DBLP:conf/eccv/QiPZHW20} designed an object-aware module to enable matching between object-centered instructions and their corresponding visual observations. EntityGraph~\cite{DBLP:conf/nips/HongOQ0G20} modeled visual-textual correlations at the entity level via a language and visual entity-relationship graph. ORIST~\cite{DBLP:conf/iccv/QiPH0H021} used object-level features instead of view-level features as input to a sequential BERT model. 
Moudgil et al.~\cite{DBLP:conf/nips/MoudgilMALB21} developed a scene- and object-aware transformer model using attention masking and view aggregation to explicitly represent objects and scenes. However, these works rely on weak supervision from environment information or coarse-grained data due to the absence of fine-grained alignment signals in visual and textual modalities.

Subsequent efforts have aimed to address the data scarcity problem. 
Zhu et al.~\cite{DBLP:conf/acl/ZhuHCDJIS20} and Hong et al.~\cite{DBLP:conf/emnlp/HongOWG20} first segmented long instructions into sub-instructions using template matching, then grounded them to visual sub-trajectories via dynamic programming and human annotation, respectively. 
Wang et al.~\cite{DBLP:conf/cvpr/0001MOBFGJWBA22} recognized entity phrases in instructions using text parsers and identified matching environmental landmarks through object detection and weak supervision from pose traces. 
However, these fine-grained cross-modal alignment signals lack precision. To construct high-quality alignments, He et al.~\cite{DBLP:conf/nips/HeHWYASW21} manually annotated the matching relationships between sub-instructions and sub-trajectories.
Additionally, Cui et al.~\cite{DBLP:conf/iccv/Cui0ZZYY23} introduced the GEL-R2R dataset, enriched with grounded entity-landmark annotations. While these efforts improved data quality, their datasets are insufficient and difficult to expand due to the high cost of human annotation.

To address the challenges outlined above, we introduce a Generative framework for Navigation Instruction annotated with Fine-grained Cross-modal Alignment (FCA-NIG). This framework is designed to generate trajectory descriptions and simultaneously annotate them with detailed cross-modal information. The FCA-NIG framework operates through six key sub-steps: trajectory chunking, landmark detection, crafted instruction construction, R2R-like instruction generation, entity selection, and sub-pair integration. Broadly, the FCA-NIG framework first segments a randomly sampled augmented trajectory into multiple sub-trajectories. These sub-trajectories are then processed sequentially through the intermediate steps to generate corresponding sub-instructions, forming several sub-instruction sub-trajectory pairs annotated with entity-landmark matches. Finally, these sub-pairs are concatenated to construct the complete instruction-trajectory pair.

During the landmark detection phase, we employ the open-world object detection model GLIP~\cite{DBLP:conf/cvpr/LiZZYLZWYZHCG22} to identify salient objects or scenes in the panoramic observation images of the sub-trajectories, which serve as reference landmarks. In the crafted instruction construction phase, we first establish a template instruction library comprising three core components: horizontal turning, vertical movement adjustments, and spatial relationships with target objects. Leveraging these templates, we construct crafted sub-instructions for each sub-trajectory based on the detected landmarks and associated action turning data. In the R2R-like instruction generation phase, we design a speaker model based on the multi-modal pre-trained language generation model OFA~\cite{DBLP:conf/icml/WangYMLBLMZZY22}, named OFA-Speaker. This model includes a panorama encoder, a vision encoder, a text encoder, a cross-modal encoder, and a text decoder, and is fine-tuned on the R2R dataset. Specifically, the OFA-Speaker uses a Transformer-based panorama encoder to extract features from the panoramic images of the sub-trajectories. These features are then concatenated with prompt words and template instructions as input to the OFA model to generate navigation sub-instructions similar to those in the R2R dataset.
In the entity selection phase, we first parse the generated sub-instructions using the NLTK~\cite{DBLP:conf/acl/Bird06} to extract the entities contained within them. We then employ the large-scale multi-modal pre-trained model CLIP~\cite{radford2021clip} to evaluate the similarity between each extracted entity and the detected visual landmarks. This process identifies the most relevant entity, which is then paired with the corresponding visual landmark to form an entity-landmark match pair.

Using the FCA-NIG framework, we generate an augmentation dataset with fine-grained cross-modal alignment annotations, named FCA-R2R. We conduct extensive experiments to evaluate the quality of instruction-trajectory pairs in the FCA-R2R dataset and the effects of two types of fine-grained alignment information: sub-instruction sub-trajectory matching and entity-landmark matching.
The evaluation involves four classic VLN agent models (SF~\cite{DBLP:conf/nips/FriedHCRAMBSKD18}, EnvDrop~\cite{DBLP:conf/naacl/TanYB19}, RecBERT~\cite{DBLP:conf/cvpr/Hong0QOG21}, and HAMT~\cite{DBLP:conf/nips/ChenGSL21}), showing that incorporating the FCA-R2R dataset significantly improves navigation performance across all agents. Further experiments using Babywalk~\cite{DBLP:conf/acl/ZhuHCDJIS20} and AFAC~\cite{DBLP:conf/icmcs/CuiHZCXYY23} methods demonstrate that adding more sub-instruction sub-trajectory matching supervision signals enhances agents' state-awareness and action decisions. Lastly, applying the GELA~\cite{DBLP:conf/iccv/Cui0ZZYY23} method highlights that sufficient entity-landmark matching supervision can further improve navigation performance, achieving notable SR and SPL scores on unseen validation and test sets. Specifically, on the unseen validation set and unseen test set, the Success Rate (SR) scores reach 71.1\% and 68.3\%, respectively, while the Success Rate weighted by Path Length (SPL) scores reach 65.5\% and 63.3\%, respectively.

To summarize, our contributions are three-fold:
\begin{compactitem}
    \item We propose a novel generative framework, FCA-NIG, designed to produce high-quality navigation instructions incorporating fine-grained cross-modal alignment annotations. 
    \item We introduce FCA-R2R, the first large-scale dataset annotated with fine-grained cross-modal alignment in the VLN domain. This dataset systematically grounds navigation instructions to both sub-trajectory segments and visual landmarks, enabling precise supervision for cross-modal alignment during navigation. 
    \item We demonstrate the effectiveness of the FCA-NIG framework and the FCA-R2R dataset through extensive experiments, showing that they significantly enhance the navigation performance of VLN agents by improving the quality of instruction-trajectory pairs and enabling more accurate action decisions.
\end{compactitem}

\section{Related work}
\subsection{Vision-Language Navigation}
The field of Vision-Language Navigation (VLN) has seen rapid advancements, driven by the integration of visual perception and language understanding for embodied decision-making. Anderson et al.~\cite{DBLP:conf/cvpr/AndersonWTB0S0G18} pioneered early work by establishing the R2R dataset based on 90 houses from the Matterport3D (MP3D) simulator~\cite{DBLP:conf/3dim/ChangDFHNSSZZ17}, enabling agents to navigate photo-realistic environments via textual instructions. Subsequent research expanded VLN to diverse domains: outdoor navigation (e.g., TOUCHDOWN~\cite{DBLP:conf/cvpr/ChenSMSA19}, AVDN~\cite{DBLP:conf/acl/FanCJZZW23}), continuous exploration (R2R-CE~\cite{krantz2021waypoint}, LN-VLN~\cite{song2024towards}), dynamic environments (ARRAMON~\cite{Kim2020ArraMonAJ}), and interactive tasks (ALFRED~\cite{ALFRED}). Instruction formats have evolved from descriptive (R2R~\cite{DBLP:conf/cvpr/AndersonWTB0S0G18}, RxR~\cite{DBLP:conf/emnlp/KuAPIB20}) to object-centric (REVERIE~\cite{DBLP:conf/cvpr/QiW0WWSH20}) and conversational (CVDN~\cite{DBLP:conf/corl/ThomasonMCZ19}, TEACh~\cite{DBLP:conf/aaai/PadmakumarTSLNG22}) paradigms.
Methodologically, VLN models have transitioned from LSTM-based architectures~\cite{DBLP:conf/iclr/MaLWAKSX19, DBLP:conf/eccv/WangWS20}) to transformer-based frameworks~\cite{DBLP:conf/eccv/MajumdarSLAPB20, DBLP:conf/iccv/GuhurTCLS21} for enhanced multimodal representation. Reinforcement learning (RCM~\cite{DBLP:conf/cvpr/WangHcGSWWZ19}, EnvDrop~\cite{DBLP:conf/naacl/TanYB19}) and imitation learning approaches have been employed to refine decision-making. Pretraining-finetuning pipelines (Prevalent~\cite{DBLP:conf/cvpr/HaoLLCG20}, HAMT~\cite{DBLP:conf/nips/ChenGSL21}) improved generalization across downstream tasks. Knowledge-augmented reasoning (CKR~\cite{DBLP:conf/cvpr/GaoC0WZW21}, KERM~\cite{DBLP:conf/cvpr/LiWYWJ23}) and backtracking policies~\cite{DBLP:conf/cvpr/MaWAXK19, DBLP:conf/cvpr/ChenGTSL22} further enhanced robustness. While these advancements addressed high-level navigation, challenges persist in leveraging fine-grained cross-modal alignment signals for precise decision-making. This paper focuses on data augmentation strategies to enrich such signals.

\subsection{VLN Data Augmentation}
Due to the limited availability of expensive human annotations, researchers have explored data augmentation methods to generate additional trajectory-instruction pairs.
Contemporary approaches have focused on data augmentation from three perspectives: image, trajectory, and environment.
Image-based techniques aim to diversify visual observations to improve robustness to environmental variations. Early work introduced environment dropout~\cite{DBLP:conf/naacl/TanYB19}, which randomly masked regions of panoramic images during training to simulate occlusions. Pathdreamer~\cite{DBLP:conf/iccv/KohLYBA21} leveraged generative hierarchical world models to synthesize complex indoor environments, though such approaches primarily enriched visual features without generating explicit trajectory annotations.
Trajectory-based methods focus on synthesizing novel paths while maintaining instruction-trajectory compatibility. Fried et al.~\cite{DBLP:conf/nips/FriedHCRAMBSKD18} pioneered this approach by sampling path viewpoints in seen environments and using a speaker model to generate instructions for sampled paths. Ossandon et al.~\cite{DBLP:conf/eccv/OssandonES22} and Wang et al.~\cite{DBLP:conf/cvpr/0001MOBFGJWBA22} extended this by designing semantically rich instructions via human templates and multimodal transformers, respectively. Liu et al.~\cite{DBLP:conf/iccv/0002ZCLGS21} proposed a mixup strategy, interpolating sub-trajectories and sub-instructions from distinct pairs to create hybrid samples. Koh et al.~\cite{DBLP:conf/aaai/KohABTWLYBA23} introduced spatial resampling, perturbing observation positions during agent training to encourage exploration of alternative paths. 
% While these methods expand trajectory diversity, they often lack fine-grained alignment signals critical for precise navigation.
To address limited coverage of environments, recent efforts scale VLN training to out-of-domain spaces. Chen et al.~\cite{DBLP:conf/eccv/ChenGTSL22} and Kamath et al.~\cite{DBLP:conf/cvpr/KamathA0KKWYBP23} utilized large-scale simulators (HM3D, Gibson) to construct navigation graphs and automatically generate instructions via speaker models. Li et al.~\cite{DBLP:conf/nips/LiB23} proposed a text-to-panorama framework, enabling synthetic environment generation through textual descriptions. Lin et al.~\cite{DBLP:conf/iccv/LinCHLTG23} exploited household video data from YouTube to extract trajectory-instruction pairs, mimicking real-world scenarios. 
Despite these innovations, most datasets remain limited to coarse-grained instruction-trajectory matching, neglecting critical sub-instruction and entity-level alignments. This paper addresses this gap by introducing a framework for scalable instruction-trajectory pairs with fine-grained cross-modal alignment signals.

\subsection{Cross-Modal Alignment in VLN}
In VLN, cross-modal alignment is fundamental to an agent’s ability to accurately interpret textual instructions and navigate visual environments. This alignment can be categorized into three hierarchical levels based on the granularity of linguistic input: instruction-level, sub-instruction-level, and entity-level alignment.
At the highest abstraction, instruction-level alignment focuses on the global correspondence between full trajectories and complete navigation instructions. Wang et al.~\cite{DBLP:conf/cvpr/WangHcGSWWZ19} pioneered the integration of cross-modal matching metrics into reinforcement learning frameworks, using the compatibility score between an agent’s trajectory and speaker-generated instructions as an intrinsic reward. Similarly, Huang et al.~\cite{DBLP:conf/iccv/HuangJMKMBI19} formalized this concept in an auxiliary task called Cross-Modal Alignment (CMA), which evaluates the semantic compatibility of instruction-trajectory pairs during training. Chen et al.~\cite{DBLP:conf/nips/ChenGSL21} and Qiao et al.~\cite{DBLP:journals/pami/QiaoQHYWW23} extended CMA to pretraining stages, leveraging large-scale in-domain and out-of-domain datasets to initialize unimodal and multimodal representations for transformer-based models.
Sub-instruction level investigates the localized mapping between sub-paths and sub-instructions. Hong et al.~\cite{DBLP:conf/emnlp/HongOWG20} and Zhu et al.~\cite{DBLP:conf/acl/ZhuHCDJIS20} introduced greedy segmentation algorithms to decompose long instructions into sequential sub-instructions, enabling agents to execute modular tasks incrementally. He et al.~\cite{DBLP:conf/nips/HeHWYASW21} manually annotated sub-instruction-sub-path alignments and incorporated focal-oriented rewards to guide local decision-making. AFAC~\cite{DBLP:conf/icmcs/CuiHZCXYY23} further advanced this domain by explicitly supervising local alignment through attention mechanisms and representation learning.
Despite progress, such methods remain limited by reliance on human annotations or heuristic decomposition strategies, which hinder scalability.
Entity-level alignment targets the correspondence between explicit landmark references in instructions and their visual counterparts. Cui et al.~\cite{DBLP:conf/iccv/Cui0ZZYY23} addressed this challenge by curating the GEL-R2R dataset, which features human-annotated entity-landmark pairs in R2R dataset. Their GELA framework demonstrated improved fine-grained alignment capabilities by conditioning models on entity-specific visual features. 
While these studies have made progress in improving cross-modal alignment, they are ineffective in addressing fine-grained cross-modal alignment due to the data scarcity problem.
In this paper, we automate the creation of large-scale, high-quality datasets with fine-grained alignment annotations by leveraging a series of advanced foundation models.

\subsection{Multimodal Text Generation}
Multimodal text generation focuses on combining visual and textual data to create coherent natural language descriptions. It spans tasks like image/video captioning and Visual Question Answering (VQA), requiring models to align modality-specific features and cross-modal relationships. Early breakthroughs, such as \emph{Show and Tell}~\cite{DBLP:conf/cvpr/VinyalsTBE15}, demonstrated end-to-end vision-language alignment using CNNs for visual feature extraction and RNNs for text generation. Xu et al.~\cite{DBLP:conf/icml/XuBKCCSZB15} later improved captions via attention mechanisms, dynamically weighting relevant image regions during decoding.
Video captioning introduced temporal coherence challenges, prompting architectures like 3D CNNs and temporal attention modules to model sequential dynamics~\cite{DBLP:conf/cvpr/AafaqALGM19}. Concurrently, VQA tasks emphasized joint reasoning over visual scenes and queries. Anderson et al.~\cite{DBLP:conf/cvpr/00010BT0GZ18} pioneered region-based attention, enabling fine-grained interaction between image segments and language tokens.
A major shift came with cross-modal pretraining, allowing models to learn universal representations across modalities without task-specific supervision. Frameworks like Unicoder-VL~\cite{DBLP:conf/aaai/LiDFGJ20} and VLP~\cite{DBLP:conf/aaai/ZhouPZHCG20} combined masked language modeling and contrastive learning to train vision-language encoders, achieving state-of-the-art results on benchmark tasks. Recent studies further unified diverse vision-and-language tasks under shared paradigms. VL-T5 and VL-BART~\cite{DBLP:conf/icml/ChoLTB21} adapted transformer architectures to integrate visual features and framed several tasks as a single-text generation problem. GIT simplified prior models by decoupling a ViT image encoder and GPT-style decoder, with GIT2 scaling up datasets and model size for better performance~\cite{DBLP:journals/tmlr/WangYHLLGLLW22}.
Wang et al.~\cite{DBLP:conf/icml/WangYMLBLMZZY22} introduced OFA, a unified framework for cross-modal and unimodal tasks (e.g., image captioning, VQA). Using an order-to-order learning approach and prompt-based task representation, OFA seamlessly integrates diverse objectives. However, while existing tasks focus on simple images,  navigation instruction generation makes complex sequential environment panoramas as input and requires precise spatial understanding absent in standard benchmarks.
In this paper, we adapt OFA for accurate instruction generation by incorporating a panoramic encoder designed to process high-resolution and context-rich scenes. By integrating this encoder into the OFA-Speaker architecture, our approach bridges the gap between general-purpose vision-language models and navigation-specific requirements, improving both scene comprehension and instruction fidelity.

\begin{figure*}\centering
\includegraphics[scale=0.85]{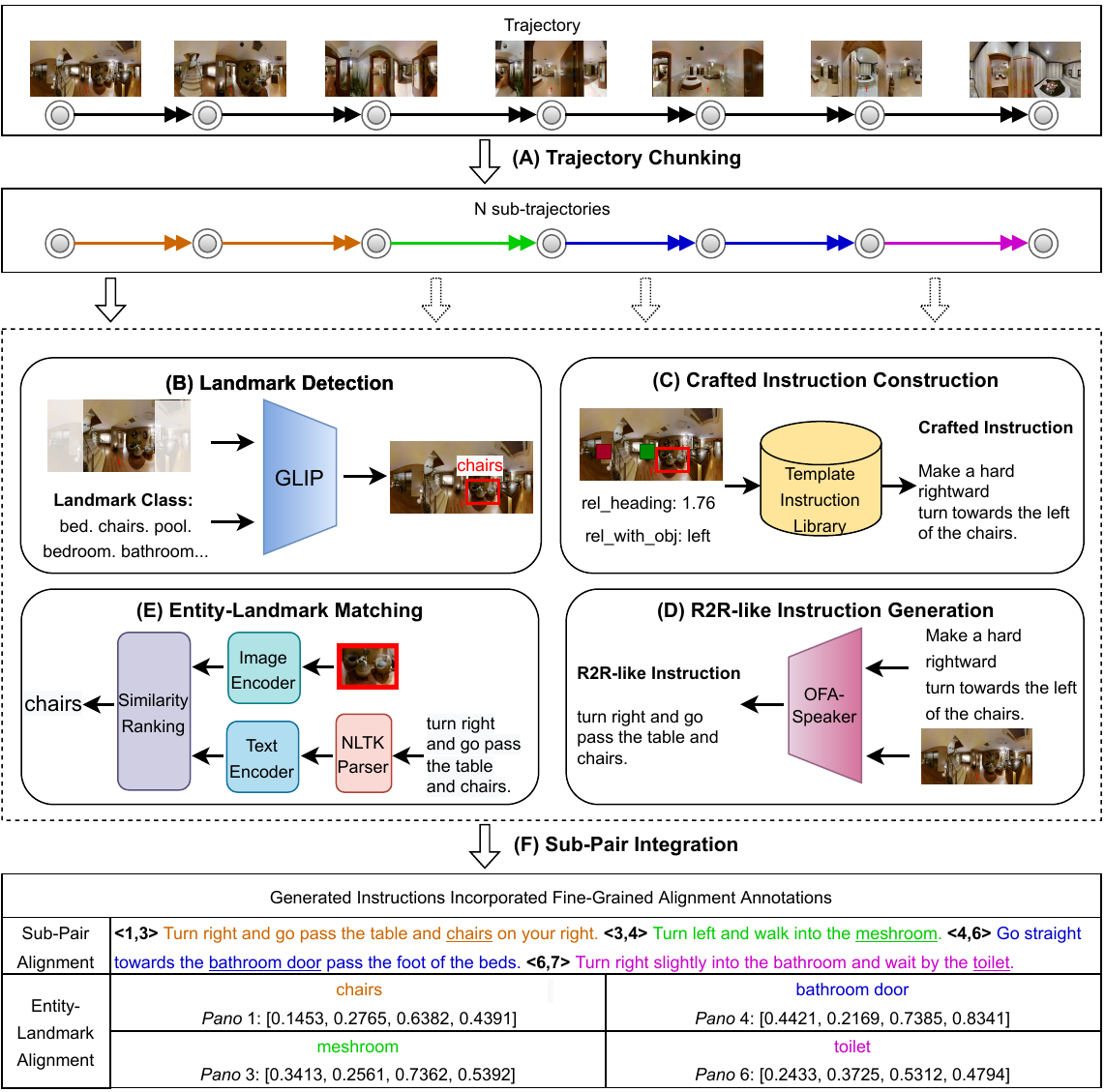}
\caption{Overview of the FCA-NIG framework. Initially, a randomly sampled augmented trajectory is segmented into multiple sub-trajectories via trajectory chunking (A). Each sub-trajectory undergoes sequential processing through intermediate steps: landmark detection (B), crafted instruction construction (C), R2R-like instruction generation (D), and entity selection (E), producing corresponding sub-instructions with entity-landmark matches. These sub-instruction sub-trajectory pairs are then integrated (F) to form the complete instruction-trajectory pair. Through this pipeline, our FCA-NIG framework generates detailed navigation instructions incorporated dual-level fine-grained alignment annotations.}
\label{fig:pipe}
\end{figure*}

\section{Preliminaries}
The previous research~\cite{DBLP:conf/nips/FriedHCRAMBSKD18} on creating synthetic VLN instructions utilized the Speaker-Follower framework. 
In this framework, the speaker produces a grounded description of navigation trajectories as their navigation instruction. 
These generating instructions could then be used to enhance the training dataset for the follower. 
The follower, in turn, makes action decisions based on human instructions and visual observations to navigate toward the target position.
In this work, we first construct a speaker-based data augmentation pipeline to generate novel trajectory-instruction pairs with fine-grained cross-modal alignments.
Then we obtain the augmented dataset to train the follower and enhance the agents' navigation performance.

\subsection{The VLN Speaker}
The goal of the VLN speaker is to generate a matching navigation instruction for a given trajectory selected from the environments. 
The speaker model typically comprises a trajectory encoder $E$ and a language decoder $D$.
As the input of the encoder, the trajectory consists of a sequence of panoramic visual observations $\mathcal{O}=\left\{O_{0}, \ldots, O_{T-1}\right\}$ along with the corresponding action $A=\left\{a_{0}, \ldots, a_{T-1}\right\}$ at each step $t$. 
Then the language decoder outputs an instruction consisting of a sequence of L words, denoted as $\mathcal{W}=\left\{w_{0}, \ldots, w_{L-1}\right\}$.
Specifically, the objective is to learn a model with parameter $\theta$ by maximizing the likelihood of the target instruction $I$ conditioned on the input $V$ and $A$: $\boldsymbol{\theta}^*=\arg \min _{\boldsymbol{\theta}} \log p(\boldsymbol{I} \mid \boldsymbol{V}, \boldsymbol{A}; \boldsymbol{\theta}) .$

\subsection{The VLN follower}
The goal of the VLN follower is to arrive at the target position from a starting pose inside a 3D environment by following the navigation instruction $I$. 
Formally, at each step $t$, the follower acquires a new panoramic visual observation $O_{t}=\left\{o_{t, i}\right\}_{i=1}^{36}$ from surrounding environment, where $o_{t, i}=\left[v_{t, i} ; a_{t, i}\right]$ consists RGB image $v_{t, i}$ and panoramic action orientation $a_{t, i}$ of ${i}$-th view. 
The follower needs to pick the next step from the action space $a_{t}=\left\{act_{t, 1}, \ldots, act_{t, n}, [STOP]\right\}$ that consists of the relative angles of $n$ navigable viewpoints to the current viewpoint and the stop action. 
The follower selects an action to move to a navigable viewpoint in a connectivity graph $G$ or stop at the current location. 
Once the follower stops, the navigation episode is completed.

\section{Methods}
\textbf{Overview.} In this study, we present the FCA-NIG framework for generating novel trajectory-instruction pairs that incorporate fine-grained cross-modal alignment annotations. 
As illustrated in the figure~\ref{fig:pipe}, the framework comprises six steps: (A) trajectory chunking, (B) landmark detection, (C) crafted instruction construction, (D) R2R-like instruction generation, (E) entity selection, and (F) sub-pair integration. 
Initially, we randomly sample a navigation trajectory from the environment and divide it into multiple sub-trajectories using step A. 
Then these sub-trajectories sequentially undergo steps B, C, D, and E to obtain their corresponding natural language navigation instruction, thus producing n sub-pairs. 
Simultaneously, the landmarks along these sub-trajectories and the corresponding entities in the sub-instructions are labeled. 
Finally, these sub-pairs are connected to obtain a trajectory-instruction pair with alignment annotations of sub-pairs and entity-landmark pairs.

\subsection{Trajectory Chunking}
To acquire sub-pair alignment signals, we partition a complete trajectory into $N$ sub-trajectories based on the agent's movement directions, denoted as  $\left\{ST_{0}, \ldots, ST_{N-1}\right\}$.
For each sub-trajectory, we generate corresponding path descriptions, referred to as sub-instructions, resulting in the set $\left\{S\mathcal{W}_{0}, \ldots, S\mathcal{W}_{n-1}\right\}$. 
The process begins by randomly sampling a trajectory within the MP3D environments, similar to R2R trajectories, typically consisting of 5 to 7 steps~\cite{DBLP:conf/cvpr/AndersonWTB0S0G18}. During this sampling, we record the turning angles associated with each navigation action. We denote $\mathbf{\psi}$ as the heading angle of the agent, and $\Delta \mathbf{\psi}$ as the change in heading angle, or turning angle, for each action.
The sampled steps are categorized into four groups based on the magnitude of $\Delta \mathbf{\psi}$: straight ($-15^{\circ} <\Delta \mathbf{\psi}< 15^{\circ}$), right ($15^{\circ}<\Delta \mathbf{\psi}<165^{\circ}$), left ($-165^{\circ}<\Delta \mathbf{\psi}<-15^{\circ}$), and backward ($165^{\circ}<\Delta \mathbf{\psi}<180^{\circ}$ or $-180^{\circ}<\Delta \mathbf{\psi}<-165^{\circ}$). 
Subsequently, successive straight actions are grouped into the same sub-trajectory, while turning actions are treated as separate sub-trajectories. This approach ensures that each sub-trajectory represents a coherent segment of the agent's movement, either moving forward or executing a significant change in direction.
Through this method, an augmented trajectory is divided into $N$ sub-trajectories, each characterized by its own movement pattern and associated sub-instruction.

\begin{table*}[tb]
\centering
% \begin{threeparttable}
% \captionsetup{width=0.5\textwidth}
\caption{
  Overview of template instructions categorized into 12 horizontal action types and 3 vertical action types, along with 3 landmark relationship categories.
  }
% \begin{center}
\resizebox{0.9\textwidth}{!}{
\begin{tabular}{lll}
			\toprule[1.5pt]
\multicolumn{1}{c}{Elements}                           & \multicolumn{1}{c}{Range} & \multicolumn{1}{c}{Template}                              \\
\midrule
\multirow{12}{*}{Horizontal Direction}       & $-15^{\circ}<\Delta \mathbf{\psi}<15^{\circ}$ & go straight                     \\
					& $15^{\circ}<\Delta \mathbf{\psi}<45^{\circ}$ & turn slightly right               \\
					& $45^{\circ}<\Delta \mathbf{\psi}<75^{\circ}$ & turn moderately right             \\
					& $75^{\circ}<\Delta \mathbf{\psi}<105^{\circ}$ & turn hardly right                 \\
					& $105^{\circ}<\Delta \mathbf{\psi}<135^{\circ}$ & turn slightly to the right rear \\
					& $135^{\circ}<\Delta \mathbf{\psi}<165^{\circ}$ & turns sharply to the right rear \\
					& $165^{\circ}<\Delta \mathbf{\psi}$ or $\Delta \mathbf{\psi}<-165^{\circ}$		&turn backwards \\
					& $-45^{\circ}<\Delta \mathbf{\psi}<-15^{\circ}$ & turn slightly left                \\
					& $-75^{\circ}<\Delta \mathbf{\psi}<-45^{\circ}$ & turn moderately left              \\
					& $-105^{\circ}<\Delta \mathbf{\psi}<-75^{\circ}$ & turn hardly left                  \\
					& $-135^{\circ}<\Delta \mathbf{\psi}<-105^{\circ}$ & turn slightly to the left rear  \\
					& $-165^{\circ}<\Delta \mathbf{\psi}<-135^{\circ}$ & turns sharply to the left rear  \\
					\midrule
\multirow{3}{*}{Vertical Direction}        & $-0.2m<\Delta \zeta<0.2m$ & towards                         \\
					& $\Delta \zeta<-0.2m$ & downwards                       \\
					& $\Delta \zeta>0.2m$ & upwards                         \\
					\midrule
\multirow{3}{*}{Position Relationship with Landmark Bounding Boxes} & $\chi_{min}<\mathbf{\psi}<\chi_{max}$ & towards                         \\
					& $\mathbf{\psi}<\chi_{min}$ & towards the left of             \\
					& $\mathbf{\psi}>\chi_{max}$ & towards the right of     \\
			\bottomrule[1.5pt]

\end{tabular}
}
% \end{center}
\label{table:tem}
% \end{threeparttable}
\end{table*}

\subsection{Landmark Detection}
To establish a robust connection between visual landmarks and textual entities, our approach integrates object detection with natural language processing. Specifically, for each sub-trajectory, we identify a prominent landmark using an advanced object detector, GLIP~\cite{DBLP:conf/cvpr/LiZZYLZWYZHCG22}, and then match this visual feature to the corresponding entity phrase derived from the generated instructions in step E.
As depicted in Step B of Figure~\ref{fig:pipe}, GLIP, a state-of-the-art pre-trained model, excels at detecting objects and grounding phrases within images. By merging object detection and phrase grounding tasks, GLIP can learn object-level, language-aware, and semantically rich visual representations, which enable it to perform exceptionally well in zero-shot scenarios in open-world environments. Given an image along with continuous text or discrete object categories as input, GLIP accurately outputs the bounding box coordinates and semantic labels of the detected objects.
To optimize GLIP's performance within our framework, we preprocess both textual and visual inputs. For the textual component, we construct a specialized library of landmark categories tailored to each MP3D environment, drawing on the GEL-R2R dataset~\cite{DBLP:conf/iccv/Cui0ZZYY23}. This category library serves as the reference for identifying landmarks through GLIP. On the visual side, we constrain the detection area to a 180-degree sector aligned with the agent's intended direction of movement, effectively halving the panorama and enhancing the precision of landmark identification. Our MP3D panoramas are captured at a resolution of 2048 x 1024, but we resize them to 1024 x 1024 to improve computational efficiency without sacrificing critical detail.
At each waypoint of a sub-trajectory, we apply GLIP to detect objects within the panoramic images, selecting the object or scene with the highest confidence score as the representative landmark. This systematic process results in a sequence of landmarks denoted as $\left\{l_{0}, \ldots, l_{N-1}\right\}$, where each $l_{i}$ corresponds to a significant point along the trajectory.

\begin{figure*}\centering
\includegraphics[scale=0.8]{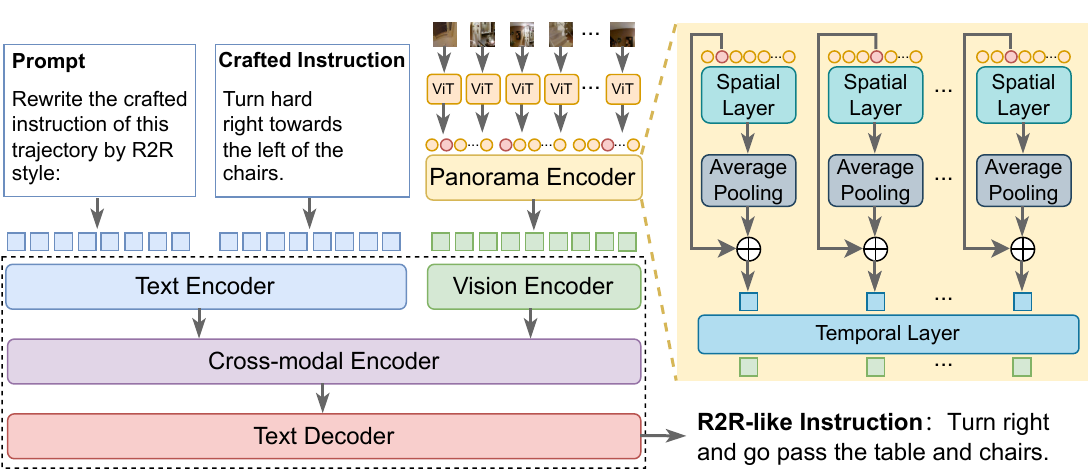}
\caption{Architecture of the OFA-Speaker model. The OFA-Speaker introduces a panorama encoder into the standard OFA model, which includes a text encoder, a vision encoder, a cross-modal encoder, and a text decoder. Specifically, the panorama encoder features spatial and temporal layers to process context-rich panoramas. The OFA-Speaker integrates prompts, crafted instructions, and a sequence of panoramas as inputs, generating R2R-like navigation instructions.}
\label{fig:model}
\end{figure*}

\subsection{Crafted Instruction Construction}
The generation of navigation instructions entails crafting a natural language description for a specified visual trajectory. 
Unlike image and video captioning tasks, which typically focus on static or sequential visual content, navigation instruction generation requires understanding the spatial relationships within a series of panoramic images. Existing multimodal generation models are yet to exhibit robust spatial reasoning capabilities, prompting our approach to begin with manually constructing detailed navigation instructions for each sub-trajectory based on the agent's vertical changes, horizontal turning angles, and the relationship between the next step's orientation and landmarks.
These manually crafted instructions are subsequently refined by a natural language generation (NLG) model into R2R-like navigation instructions, ensuring they are both precise and user-friendly.
As illustrated in the table~\ref{table:tem}, we refine the classification of right and left turns in step A, with each category corresponding to a 30-degree angular segment. This finer granularity allows for more accurate descriptions of the agent's movements. In conjunction with the actions of going straight and turning back, we classify the agent's horizontal movements into 12 distinct categories.
Let $\zeta$ denote the vertical coordinates of the agent, while $\chi_{min}$ and $\chi_{max}$ represent the heading angles of the agent aligned with the left and right boundaries of the landmark's boundary box, respectively. Vertical movements are categorized into three types: horizontal, upward, and downward. Movements with $-0.2m<\Delta \zeta<0.2m$ are deemed horizontal, those with $\Delta \zeta>0.2m$ are classified as upward, and those with $\Delta \zeta<-0.2m$ as downward. 
The orientation of the next step relative to landmarks is categorized into three types: "towards" ($\chi_{min}<\mathbf{\psi}<\chi_{max}$), "towards the left of" ($\mathbf{\psi}<\chi_{min}$), and "towards the right of" ($\mathbf{\psi}>\chi_{max}$). 
As shown in Table~\ref {table:tem}, based on the 12 horizontal action categories and 3 vertical action categories, combined with the 3 landmark relationship categories, we construct a template instruction library consisting of 108 unique template instructions. For each sub-trajectory, we select the most appropriate template from this library and combine it with the detected landmarks to form a crafted navigation instruction.
As depicted in step C of the Figure~\ref{fig:pipe}, the red dot represents the current orientation of the agent, while the green dot indicates the next step's orientation. Based on these inputs, we generate a crafted instruction such as: "Turn slightly to the left and walk towards the left side of the chair."

\subsection{R2R-like Instruction Generation}

Recent progress in multimodal pretraining models has significantly advanced language generation capabilities within the vision-language domain. However, the Speaker models have historically struggled due to their insufficient ability to align vision and language modalities effectively~\cite{DBLP:conf/nips/FriedHCRAMBSKD18, DBLP:journals/eaai/WangLHLYCC24, DBLP:conf/cvpr/0001MOBFGJWBA22}. To address this limitation, we develop a VLN Speaker model based on the large-scale multimodal pretraining model OFA~\cite{DBLP:conf/icml/WangYMLBLMZZY22}, which we designate as OFA-Speaker. By adopting an order-to-order learning framework and a prompt-based task representation, OFA seamlessly integrates various unimodal and multimodal tasks, such as image classification, text filling, image captioning, and VQA.
Despite these advancements, directly applying the OFA model to navigation instruction generation poses challenges. The existing tasks primarily involve simple natural images, which are far less complex than the sequential environment panoramas required for navigation. To enhance the model's understanding of complex scenes, we introduce a panoramic image encoder into the OFA-Speaker model, as shown in figure~\ref{fig:model}. Additionally, we design a specific prompt—"Manual instructions to rewrite this path in the style of R2R instructions"—to expand OFA's application to the new downstream task of R2R-like instruction generation.

\textbf{Model Architecture.} 
The classical speaker models~\cite{DBLP:conf/nips/FriedHCRAMBSKD18, DBLP:journals/eaai/WangLHLYCC24} typically process a sequence of view images as input but struggles to effectively utilize scene information due to its limited capacity for understanding spatial relationships. In contrast, the MARKY-mT5 speaker model~\cite{DBLP:conf/cvpr/0001MOBFGJWBA22} modifies the input to include prompt words and carefully crafted navigation instructions, with visual inputs restricted to landmarks. Although this approach enables the model to focus on essential navigation elements, it sacrifices some environmental context crucial for generating detailed instructions.
Our proposed OFA-Speaker integrates prompts, crafted instructions, and a sequence of panoramic images as inputs. For each viewpoint observation \(O_t\), we embed its constituent view images using the pre-trained CLIP-ViT model~\cite{radford2021clip}, denoted as \(V^t = \{v^t_0, v^t_1, \ldots, v^t_{35}\}\). These embeddings are then fed into a panorama encoder, which consists of two Transformer layers: a spatial layer that learns intra-panorama spatial relationships, and a temporal layer that captures inter-panorama temporal dynamics. An average pooling operation is applied between these layers to derive panorama embeddings, which are subsequently combined with oriented view image features via residual connections. This process yields a panorama representation defined as \(\{P^0, P^1, \ldots\}\).
Following visual processing, the panorama feature sequence is concatenated with prompt and template instructions before being input into the OFA encoder. The decoder then generates R2R-style navigation instructions, represented as \(\mathcal{W}' = \{w'_0, w'_1, \ldots\}\).

\textbf{Training and Inference.}
We train the OFA-Speaker model in two phases. In the first phase, we freeze the weights of the pre-trained OFA model and train the remaining parameters. In the second phase, we fine-tune the entire OFA-Speaker model. The training data are sourced from the R2R training set. We construct crafted instructions for the R2R training set paths using entity-landmark pairs from human-annotated GEL-R2R, pairing them with ground-truth instructions to form training data. During inference, the entity classes and landmarks used in constructing these crafted instructions are detected using GLIP.
To address the issue of textual repetition in generated instructions, we adapt the SimCTG approach~\cite{su2023contrastive}, which consists of two components: contrastive training algorithms during training and contrastive search decoding methods during inference. Repetitive text generation arises from the anisotropic nature of the token representations produced by the model, where the vector representations of different tokens within a sentence are overly similar, leading to repeated token generation. Contrastive training aims to adjust the model's representation space to maximize the distinctiveness of different tokens within the same sentence. This is achieved by integrating a contrastive loss function with the traditional maximum likelihood estimation (MLE) loss function:
\begin{equation}
	\mathcal{L}_\text{MLE} = -\frac{1}{|x|} \sum_{i=1}^{|x|} \log p_\theta(x_i | x_{<i}),
\end{equation}
\begin{equation}
\begin{aligned}
	\mathcal{L}_\text{CL} = &\frac{1}{|x|(|x| - 1)} \sum_{i=1}^{|x|} \sum_{j=1, j \neq i}^{|x|} \\ &\max\{0, \rho - c(r_{x_i}, r_{x_i}) + c(r_{x_i}, r_{x_j})\},
\end{aligned}
\end{equation}
\begin{equation}
\mathcal{L}_\text{SimCTG} = \mathcal{L}_\text{MLE} + \mathcal{L}_\text{CL},
\end{equation}
where $r$ is the representation of the model-generated token $x$, $c$ calculates the cosine similarity between token representations, and $\rho \in [ -1,1]$ is a predefined range.
In contrastive search, the aim is to enhance the diversity of generated text while preserving semantic consistency with preceding text:
\begin{equation}
\begin{aligned}
x_t=&\underset{u \in U^{(k)}}{\arg \max }\{(1-\alpha) \times \underbrace{p_\theta\left(u \mid x_{<t}\right)}_{\text {model confidence}}- \\ &\alpha \times \underbrace{\left(\max \left\{c\left(r_u, r_{x_j}\right): 1 \leq j \leq t-1\right\}\right)}_{\text {degeneration penalty}}\},
\end{aligned}
\end{equation}
where $U^{(k)}$ denotes the most likely candidate set of the top $K$ predicted by the model, $p_\theta\left(u \mid x_{<t}\right)$ is the probability of the candidate $u$ given the previous context $x$, and $alpha$ is a hyperparameter balancing model confidence and degeneration penalties. This decoding strategy prefers tokens with low similarity to prior representations, thereby avoiding model degradation and ensuring textual coherence.

\subsection{Entity Selection}
Given that the visual input consists of panoramic observations containing numerous objects, the navigation instructions generated by OFA-Speaker may include multiple entity phrases, denoted as \(\left\{e_{i,0}, e_{i,1}, \ldots\right\}\). To obtain entity-level cross-modal matching annotations for each sub-pair in Step B, we need to select the most relevant entity corresponding to the landmark on the sub-trajectory.
We begin by using the NLTK text parser~\cite{DBLP:conf/acl/Bird06} to extract entities from the generated sub-instructions. These extracted entities serve as candidate landmark entities for each sub-instruction. This step ensures that we have a comprehensive list of potential landmarks mentioned in the text.
To select the most appropriate entity, we employ a multimodal pretraining model, CLIP~\cite{radford2021clip}, which uses a dual-encoder structure to map both text and image data into a shared semantic space. CLIP is pretrained using large-scale contrastive learning, enabling it to exhibit strong zero-shot learning capabilities, generalization, and transferability. This robust foundation makes CLIP particularly well-suited for cross-modal understanding and generation tasks.
Specifically, in this step, the environment landmark \(l_{i}\) is fed into the image encoder of the CLIP model to extract image features. The candidate entities are input into the text encoder of the CLIP model to obtain text features. The cosine similarity between the text features and image features is then calculated and ranked. The entity with the highest cosine similarity score is selected as the corresponding entity \(e_{i}\) for the reference landmark \(l_{i}\).
By leveraging the powerful capabilities of CLIP, we can effectively bridge the gap between visual and textual information, thereby enhancing the accuracy and reliability of the entity-level cross-modal matching annotations.

\subsection{Sub-Pair Integration}
Through the above steps, we derive sub-instructions for each sub-trajectory and extract entity-landmark alignment. These sub-instructions are then connected to form a complete instruction. During this process, the positions of marker entities within the sub-instructions are adjusted to correspond to their positions within the full instruction. As a result, the system generates a complete instruction-trajectory pair with fine-grained cross-modal matching, enabling precise alignment between visual and textual modalities.

\section{Experiments and results}

\subsection{Experimental Setup}

\subsubsection{Dataset}

We train the OFA-Speaker using training pairs constructed from the R2R dataset~\cite{DBLP:conf/cvpr/AndersonWTB0S0G18} and evaluate the effectiveness of the augmentation dataset generated by the OFA-Speaker for the follower agent. 
The R2R dataset, based on 90 photo-realistic indoor environments within the Matterport3D simulator, comprises 7,189 shortest-path trajectories derived from navigation graphs. 
Each trajectory includes between 5 to 7 steps, with an average physical path length of 10 meters, and is annotated with three human-generated instructions, resulting in a total of 21,500 instructions. 
On average, each instruction contains approximately 29 words.
The dataset is divided into four subsets: training (61 environments), validation seen (61 environments), validation unseen (11 environments), and test unseen (18 environments). 
Notably, the validation seen split comprises environments identical to those in the training set, while the validation unseen and test splits contain distinct environments. 
% The inclusion of trajectories in unseen environments enables the assessment of the agent’s generalization ability.
The inclusion of these unseen trajectories is crucial for assessing the model’s adaptability and performance in novel settings.

\begin{table*}[tb]
\centering
% \begin{center}
% \captionsetup{width=0.85\textwidth}
  \caption{
  Comparative Statistics of VLN Datasets with Fine-Grained Cross-Modal Matching Annotations.
  }
  \resizebox{0.85\textwidth}{!}{
\begin{tabular}{lccccc}
			\toprule[1.25pt]
			\multicolumn{1}{c}{Datasets} &
			\multicolumn{1}{c}{Sub-instruction level} &
			\multicolumn{1}{c}{Entity level} &
			\multicolumn{1}{c}{Trajectory} &
			\multicolumn{1}{c}{Instruction} &
			Fine-grained pairs \\
			\midrule
			FG-R2R~\cite{DBLP:conf/emnlp/HongOWG20}       & \ding{51}  &  & 5,798    & 17,409   & 63,633   \\
			% GSP-R2R      & \chmark &  & \numprint{5798}    & \numprint{17409}     & \numprint{46958}   \\
			Landmark-RxR~\cite{DBLP:conf/nips/HeHWYASW21} & \ding{51} &  & 11,321  & 26,464  & 166,740  \\
			GEL-R2R~\cite{DBLP:conf/iccv/Cui0ZZYY23}      &  & \ding{51} & 5,798 & 17,409  & 150,183  \\
			Marky~\cite{DBLP:conf/cvpr/WangLSGW22}        &  & \ding{51} & 33,889  & 101,669 & 970,762  \\
			\midrule
			\multirow{2}{*}{FCA-R2R(ours)} 
			& \ding{51} & & \textbf{178,270} & \textbf{534,810} &\textbf{2,519,919} \\
			&  & \ding{51} & \textbf{178,270} & \textbf{534,810}    & \textbf{5,240,981} \\
			\bottomrule[1.25pt]
\end{tabular}
}
% \end{center}

  \label{table:data}
\end{table*}

\subsubsection{Evaluation Metrics}

We evaluate the quality of our generated instructions using automatic language metrics, reporting results based on BLEU~\cite{DBLP:conf/acl/PapineniRWZ02}, METEOR~\cite{DBLP:conf/wmt/DenkowskiL14}, ROUGE~\cite{lin-2004-rouge}, and CIDEr~\cite{vedantam2015cider}, consistent with previous studies~\cite{DBLP:conf/nips/FriedHCRAMBSKD18, DBLP:conf/cvpr/0001MOBFGJWBA22, DBLP:conf/cvpr/WangWSY23}. Additionally, to assess the efficacy of our augmentation dataset in enhancing the follower agents' performance, we employ four standard navigation performance metrics:
1) Trajectory Length (TL): Measures the total distance traveled along the navigation path.
2) Navigation Error (NE): Defines the shortest distance between the agent's final position and the target viewpoint.
3)Success Rate (SR): Indicates the proportion of trials where the agent reaches within a 3-meter radius of the target viewpoint.
4) Success Rate weighted by Path Length (SPL): Balances the success rate against the efficiency of the route taken, providing a comprehensive evaluation metric.
Particularly, SPL is endorsed as the principal metric due to its balanced consideration of both accuracy and efficiency of navigation. This dual approach ensures a thorough evaluation of both linguistic quality and navigational effectiveness, offering a holistic view of the system's capabilities.

\subsubsection{Implementation Details}

During the navigation instruction generation phase, we utilize the OFA-base pre-trained model~\cite{DBLP:conf/icml/WangYMLBLMZZY22} as our backbone model, adhering to its associated hyperparameters. Specifically, fine-tuning is performed over 200k iterations on four NVIDIA Tesla A100 GPUs, employing AdamW optimizer with a learning rate of \(2 \times 10^{-5}\) and a batch size of 64 per GPU. For the pre-training of the follower agent model, two NVIDIA RTX 3090 GPUs are used for 200k iterations, using AdamW optimizer with a learning rate of \(5 \times 10^{-5}\) and a batch size of 64 per GPU. In the subsequent fine-tuning for downstream tasks, we conduct iterative training on a single NVIDIA RTX 3090 GPU using a learning rate of \(10^{-5}\) and a reduced batch size of 16.

\begin{table*}[tb]
\centering
  \caption{
  Navigation performance improvements with FCA-R2R dataset. 
  }
  % \begin{center}
  % \resizebox{0.72\textwidth}{!}{
\begin{tabular}{lccc>{\columncolor[gray]{0.9}}cccc>{\columncolor[gray]{0.9}}c}
			\toprule[1.25pt]
			\multicolumn{1}{c}{\multirow{2}{*}{Methods}} & \multicolumn{4}{c}{Validation Seen}   & \multicolumn{4}{c}{Validation Unseen} \\
			\cmidrule{2-9} 
			\multicolumn{1}{c}{}                       & TL   & NE $\downarrow$   & SR $\uparrow$  & SPL $\uparrow$ & TL   & NE $\downarrow$   & SR $\uparrow$  & SPL $\uparrow$ \\
			\midrule
			SF~\cite{DBLP:conf/cvpr/AndersonWTB0S0G18}              & --       & \textbf{3.36} & \textbf{66.0} & --    & --       & 6.62 & 35.0 & --    \\
			SF~\cite{DBLP:conf/cvpr/AndersonWTB0S0G18}+FCA-R2R            & --       & 3.72 & 64.3 & --   & --       & \textbf{5.51} & \textbf{45.0} & --    \\
			\midrule
			EnvDrop~\cite{DBLP:conf/naacl/TanYB19}         & 11.00   & 3.99 & 62.1 & 59.0 & 10.70   & 5.22 & 52.2 & 48.0 \\
			EnvDrop~\cite{DBLP:conf/naacl/TanYB19}+FCA-R2R            & 13.42   & \textbf{3.51} & \textbf{68.6} & \textbf{63.0} & 16.67   & \textbf{4.83} & \textbf{55.8} & \textbf{49.6} \\
			\midrule
			RecBERT~\cite{DBLP:conf/cvpr/Hong0QOG21}         & 10.79   & 3.11 & 71.1 & 67.2 & 11.86   & 4.29 & 58.7 & 53.4 \\
			RecBERT~\cite{DBLP:conf/cvpr/Hong0QOG21} +FCA-R2R            & 10.98 & \textbf{2.76} & \textbf{74.3} & \textbf{70.0} & 11.67 & \textbf{4.06} & \textbf{61.6} & \textbf{56.0} \\
			\midrule
			HAMT~\cite{DBLP:conf/nips/ChenGSL21}            & 11.15   & 2.51 & 75.0 & 71.7 & 11.46   & 3.62 & 65.7 & 60.9 \\
			HAMT~\cite{DBLP:conf/nips/ChenGSL21}+FCA-R2R           & 10.80   & \textbf{2.30} & \textbf{77.8} & \textbf{74.5} & 11.68   & \textbf{3.50} & \textbf{67.2} & \textbf{61.5} \\
			\bottomrule[1.25pt]
\end{tabular}
% }
% \end{center}

  \label{table:pair}
\end{table*}

\subsection{Dataset Statistics and Comparative Analysis}

Table~\ref{table:data} presents the data statistics for trajectories, instructions, and fine-grained pairs in various VLN datasets featuring fine-grained cross-modal matching annotations. The datasets analyzed include FG-R2R~\cite{DBLP:conf/emnlp/HongOWG20}, Landmark RxR~\cite{DBLP:conf/nips/HeHWYASW21}, Marky~\cite{DBLP:conf/cvpr/WangLSGW22}, GEL-R2R~\cite{DBLP:conf/iccv/Cui0ZZYY23}, and the newly proposed FCA-R2R.
FG-R2R and Landmark RxR datasets contain only sub-pairs, whereas GEL-R2R and Marky datasets consist exclusively of entity-landmark pairs. In contrast, the FCA-R2R dataset uniquely offers both sub-pairs and entity-landmark pairs. FG-R2R and GEL-R2R are constructed using the R2R training set, seen validation set, and unseen validation set, resulting in consistent trajectory and instruction counts of 5,798 and 17,409, respectively. Landmark RxR is derived from the English portion of the RxR dataset, with 11,321 trajectories and 26,464 instructions. Marky, based on the RxR training and validation sets, comprises 33,889 trajectories and 101,669 instructions. The FCA-R2R dataset samples all 5-7 step trajectories within the Matterport3D environment, generating six instructions per trajectory, totaling 178,270 trajectories and 534,810 instructions. This significantly surpasses other datasets in terms of both trajectory and instruction volume, highlighting its substantial advantage in data scale.
Regarding fine-grained matching at the sub-instruction level, FG-R2R and Landmark RxR contain 63,633 and 166,740 sub-pairs, respectively, while FCA-R2R includes 2,519,919 sub-pairs. For entity-level fine-grained matching, GEL-R2R, Marky, and FCA-R2R have 150,183, 970,762, and 5,240,981 entity-landmark pairs, respectively. These figures demonstrate that FCA-R2R's sub-pairs and entity-landmark pairs reach into the millions, an order of magnitude larger than other datasets, underscoring its significant scale advantage.
In conclusion, the FCA-R2R dataset, with its extensive coverage of sub-instruction and entity-level fine-grained information and large-scale data, provides rich and valuable resources for advancing research in the VLN domain.

\subsection{Effectiveness of Instruction-Trajectory Pair}

This section evaluates the quality of instruction-trajectory pairs in the FCA-R2R dataset using four representative VLN models: SF~\cite{DBLP:conf/cvpr/AndersonWTB0S0G18}, EnvDrop~\cite{DBLP:conf/naacl/TanYB19}, RecBERT~\cite{DBLP:conf/cvpr/Hong0QOG21}, and HAMT~\cite{DBLP:conf/nips/ChenGSL21}. SF and EnvDrop are based on LSTM architectures, while RecBERT and HAMT utilize Transformer architectures. All four models were pretrained or fine-tuned using enhanced datasets: SF with SF-R2R, EnvDrop with EnvDrop-R2R, and RecBERT and HAMT with Prevalent-R2R. In our experiments, the original augmented datasets were replaced with the FCA-R2R dataset. 
The experimental results, presented in Table \ref{table:pair}, demonstrate improvements across most navigation metrics for all four models after training with the FCA-R2R dataset. Specifically, the SF model exhibits a significant improvement in SR by 10\% on the unseen validation set. EnvDrop shows SPL increases of 4\% and 1.6\% on the unseen validation set. RecBERT improves SPL by 2.8\% and 2.4\%, while HAMT achieves SPL improvements of 2.8\% and 0.6\% on the unseen validation set. These findings suggest that the instruction-path pairs in the FCA-R2R dataset are of higher quality.
Moreover, the performance gains for RecBERT and HAMT are notably lower than those for SF and EnvDrop. This discrepancy may stem from the fact that SF and EnvDrop were not pretrained and thus performed less effectively in multimodal feature representation compared to the pretrained RecBERT and HAMT models. Consequently, lower-quality datasets may hinder effective information extraction and limit navigation performance improvements for these models. In contrast, high-quality data like FCA-R2R enables more effective feature learning, leading to significant enhancements in navigation capabilities.

\begin{table*}[tb]
\centering
  \caption{
  Navigation performance improvements with sub-instruction sub-trajectory alignment in FCA-R2R dataset. 
  }
% \begin{center}
  \resizebox{0.72\textwidth}{!}{
\begin{tabular}{lccc>{\columncolor[gray]{0.9}}cccc>{\columncolor[gray]{0.9}}c}
			\toprule[1.25pt]
			\multicolumn{1}{c}{\multirow{2}{*}{Methods}} & \multicolumn{4}{c}{Validation Seen}   & \multicolumn{4}{c}{Validation Unseen} \\
			\cmidrule{2-9} 
			\multicolumn{1}{c}{}                       & TL   & NE $\downarrow$   & SR $\uparrow$  & SPL $\uparrow$ & TL   & NE $\downarrow$   & SR $\uparrow$  & SPL $\uparrow$ \\
			\midrule
			Babywalk~\cite{DBLP:conf/acl/ZhuHCDJIS20}  & --      & --     & --     & --     & 10.2  & 5.9   & 43.8  & 39.6  \\
			Babywalk~\cite{DBLP:conf/acl/ZhuHCDJIS20}+FCA-R2R     & 10.16 & 4.75 & 0.55 & 52.6 & 9.46 & \textbf{5.68} & \textbf{46.1} & \textbf{42.9} \\
			\midrule
			  AFAC~\cite{DBLP:conf/icmcs/CuiHZCXYY23}     & 10.92  & \textbf{2.40}  & 77.3 & 73.9 & 11.8 & 3.56  & 67.0 & 61.7 \\
			  AFAC~\cite{DBLP:conf/icmcs/CuiHZCXYY23}+FCA-R2R    & 11.01  & 2.48  & \textbf{78.0} & \textbf{74.5} & 12.0 & \textbf{3.35}  & \textbf{68.0} & \textbf{62.7} \\
			\bottomrule[1.25pt]
\end{tabular}
}
% \end{center}

  \label{table:sub}
\end{table*}

\subsection{Effectiveness of Sub-Pair Alignment}

This section evaluates the sub-pair alignment annotations in the FAC-R2R dataset using two distinct methods. The first method, Babywalk~\cite{DBLP:conf/acl/ZhuHCDJIS20}, initially constructs sub-instructions and sub-trajectory matches using heuristic algorithms on the R2R and SF-R2R datasets. It then trains agents under imitation learning and curriculum reinforcement learning frameworks, allowing progressive learning from shorter to longer paths, thereby enhancing final navigation performance. The second method, AFAC~\cite{DBLP:conf/icmcs/CuiHZCXYY23}, improves agent decision-making by incorporating auxiliary sub-pair alignment constraints. 
We applied the sub-pair matching information from the FCA-R2R dataset to train agents using both methods. The experimental results are summarized in Table~\ref{table:sub}. The results indicate that both methods improved all navigation evaluation metrics after training with the FCA-R2R dataset.  Specifically, the Babywalk method achieves SR and SPL increases of 2.3\% and 3.3\%, respectively, on the unseen validation set. Meanwhile, the AFAC method improves SR and SPL by 0.7\% and 0.6\%, respectively, on the seen validation set, and by 1.0\% for both metrics on the unseen validation set.
These findings indicate that incorporating additional sub-pair matching supervision into the training data facilitates more effective state-aware learning for agents, enabling them to make more accurate navigational decisions. This enhanced supervision significantly benefits agent performance, particularly in complex or unseen environments.

\begin{table}[tb]
\centering
  \caption{
  Navigation performance improvements with entity-landmark alignment in FCA-R2R dataset. 
  }
% \begin{center}
  \resizebox{0.5\textwidth}{!}{
\begin{tabular}{llccc>{\columncolor[gray]{0.9}}c}
			\toprule[1.25pt]
			\multicolumn{1}{c}{Methods} & \multicolumn{1}{c}{Datasets}  & TL       & NE$\downarrow$      & SR$\uparrow$       & SPL$\uparrow$     \\
			\midrule
			\multirow{3}{*}{GELA~\cite{DBLP:conf/iccv/Cui0ZZYY23}}  & val seen   & 11.19    & 2.39    & 76.4    & 72.9   \\
			& val unseen  & 11.73    & 3.11    & \textbf{71.1}    & 65.0   \\
			& val test  & 12.99    & 3.58    & 67.4    & 62.1   \\
			\midrule
			\multirow{3}{*}{GELA~\cite{DBLP:conf/iccv/Cui0ZZYY23}+FCA-R2R}  & val seen     & 11.00    & \textbf{2.31}    & \textbf{76.5}    & \textbf{73.5}   \\
			& val unseen    & 11.46    & \textbf{3.09}    & \textbf{71.1}    & \textbf{65.5}  \\
			& val test    & 12.65    & \textbf{3.57}    & \textbf{68.3}    & \textbf{63.3} \\
			\bottomrule[1.25pt]
\end{tabular}
}
% \end{center}

  \label{table:el}
\end{table}

\subsection{Effectiveness of Entity-Landmark Alignment}

This section evaluates the entity-landmark alignment annotations in the FAC-R2R dataset using the GELA method~\cite{DBLP:conf/iccv/Cui0ZZYY23}, which introduces a grounded entity-landmark adaptive pre-training framework to improve fine-grained semantic representation alignment based on human-annotated entity-landmark pairs. We utilized entity-landmark pairs from the GEL-R2R and FCA-R2R datasets for adaptive pre-training, followed by fine-tuning on the R2R dataset. The experimental results are presented in Table~\ref{table:el}.
The results indicate that the GELA method improves most navigation evaluation metrics after training with the FCA-R2R dataset. Specifically, on the seen validation set, SR and SPL improve by 0.1\% and 0.6\%, respectively. On the unseen validation set, SR remains consistent, while SPL improved by 0.5\%. On the unseen test set, SR and SPL increase by 0.9\% and 1.2\%, respectively.
These findings suggest that incorporating additional entity-landmark matching information during the adaptive pre-training stage enables the agent to learn more effective cross-modal representations. This enhanced representation learning promotes better understanding of instructions and the environment, thereby improving overall navigation performance.

\subsection{Ablation Study}
This section performs an ablation study on the quality of instructions in the FCA-R2R augmented dataset, focusing on three aspects of the OFA-Speaker model: input data, fine-tuning modules, and decoding methods. By systematically analyzing these components, we aim to identify how each aspect contributes to the overall performance of navigation instructions generated by the OFA-Speaker model. Specifically, we examine the impact of different types of input data on instruction quality, evaluate various fine-tuning strategies for optimizing model parameters, and compare different decoding approaches to enhance the clarity and accuracy of generated instructions.

\begin{table*}[tb]
\centering
  \caption{
   Ablation study results of the OFA-Speaker model input data. 
  }
% \begin{center}
  \resizebox{\textwidth}{!}{
\begin{tabular}{ccccc|cccc|cccc}
			\toprule[1.25pt]
			\multirow{2}{*}{Model} & \multicolumn{4}{c}{input data}    & \multicolumn{4}{c}{val seen}  & \multicolumn{4}{c}{val seen}    \\
			\cmidrule{2-13}
			& \makecell{crafted\\instruction} & prompt & pano & \makecell{pano\\encoder} & Bleu4 $\uparrow$  & Meteor $\uparrow$  & Rouge $\uparrow$  & CIDEr $\uparrow$  & Bleu4 $\uparrow$  & Meteor $\uparrow$  & Rouge $\uparrow$  & CIDEr $\uparrow$ \\
			\midrule
			Marky~\cite{DBLP:conf/cvpr/0001MOBFGJWBA22}          & \chmark    & \chmark   &     &        & 6.2   &  --      &  --     & 8.4   & 5.8   &   -   & -   & 7.5  \\
			\midrule
			0                   & \chmark    &     &     &        & 10.89 & 15.8   & 31.98 & 6.23  & 9.62  & 15.38  & 31.56 & 4.77 \\
			1                   & \chmark    & \chmark   &     &        & 11.58 & 16.74  & 33.32 & 5.78  & 10.97 & 15.97  & 32.39 & 4.56  \\
			2                   & \chmark    &     & \chmark   &        & 11.9  & 16.47  & 33.59 & 9.2    & 10.83 & 15.62  & 33.34 & 8.9   \\
			3                   & \chmark    & \chmark   & \chmark   &        & 12.42 & 16.71  & 34.62 & 11.02   & 12.31 & 16.47  & 34.79 & 10.51  \\
			4                   & \chmark    & \chmark   & \chmark   & \chmark      & \textbf{13.55} & \textbf{17.51}  & \textbf{35.6}  & \textbf{12.97}  & \textbf{12.99} & \textbf{17.12}  & \textbf{35.43}  & \textbf{11.79} \\
			\bottomrule[1.25pt]
\end{tabular}
}
% \end{center}

  \label{table:input}
\end{table*}

\begin{table*}[tb]
\centering
  \caption{
  Ablation study results of the OFA-Speaker model fine-tuning module. 
  }
% \begin{center}
  \resizebox{0.95\textwidth}{!}{
\begin{tabular}{ccccccccccc}
			\toprule[1.25pt]
\multirow{2}{*}{model} & \multirow{2}{*}{encoder} & \multirow{2}{*}{decoder} & \multicolumn{4}{c}{val seen}  & \multicolumn{4}{c}{val unseen}    \\
\cmidrule{4-11}
&                      &                      & Bleu4 $\uparrow$  & Meteor $\uparrow$  & Rouge $\uparrow$  & CIDEr $\uparrow$  & Bleu4 $\uparrow$  & Meteor $\uparrow$  & Rouge $\uparrow$  & CIDEr $\uparrow$  \\
\midrule
5                   & \Xmark                    & \Xmark                   & 6     & 14.89  & 30.5  & 1.02  & 5.5   & 14.5   & 30.04 & 0.77  \\
6                   & \chmark                   & \Xmark                   & 7.95  & 14.85  & 29.58 & 3.46   & 6.75  & 14.44  & 29.24 & 3.25 \\
7                   & \Xmark                    & \chmark                    & 11.2  & 17.62  & 33.37 & 6.21  & 10.67 & 17.53  & 33    & 6.2   \\
4                   & \chmark                    & \chmark       & \textbf{13.55} & \textbf{17.51}  & \textbf{35.6}  & \textbf{12.97}  & \textbf{12.99} & \textbf{17.12}  & \textbf{35.43}  & \textbf{11.79} \\
			\bottomrule[1.25pt]
\end{tabular}
}
% \end{center}

  \label{table:ft}
\end{table*}

\begin{table*}[tb]
\centering
  \caption{
  Ablation study results of the OFA-Speaker model decode method.  
  }
% \begin{center}
  \resizebox{0.7\textwidth}{!}{
\begin{tabular}{ccccccc}
			\toprule[1.25pt]
	model  & decode methods & split & Bleu4 $\uparrow$  & Meteor $\uparrow$  & Rouge $\uparrow$  & CIDEr $\uparrow$    \\
	\midrule
	\multirow{2}{*}{8}  & \multirow{2}{*}{Greedy} & val seen          & 12.7  & 17.6  & 35.06 & 11.83 \\
	&                         & val unseen                 & 11.72 & 17.01 & 34.46 & 9.88  \\
	\midrule
	\multirow{2}{*}{4} & \multirow{2}{*}{SimCTG} & val seen     & \textbf{13.55} & \textbf{17.51}  & \textbf{35.6}  & \textbf{12.97} \\
	&                         & val unseen                & \textbf{12.99} & \textbf{17.12}  & \textbf{35.43}  & \textbf{11.79}  \\
			\bottomrule[1.25pt]
\end{tabular}
}
% \end{center}

  \label{table:dec}
\end{table*}

\subsubsection{Input Data}

We begin by examining the impact of input information on the OFA-Speaker model's performance. Table \ref{table:input} presents results on both seen and unseen validation sets. Our approach contrasts with Wang et al.~\cite{DBLP:conf/cvpr/0001MOBFGJWBA22}, who utilize the mT5 model to generate the Marky dataset. Their method involves detecting landmarks using an object detection model, constructing template instructions based on actions and landmarks, and generating natural language navigation instructions via the mT5 model. However, the Marky dataset lacks simultaneous sub-pair and entity-landmark alignment information, and the mT5 model primarily focuses on natural language processing tasks with limited multimodal capabilities. 
The input to the Marky-mT5 model consists of manual instructions and prompt words. In contrast, our Model 0 uses only manual instructions as input, while Model 1 incorporates both manual instructions and prompt words. Both models significantly improve BLEU-4 scores compared to the Marky dataset, achieving increases of 4.69 and 5.38 points on the seen validation set and 3.78 and 5.17 points on the unseen validation set. However, CIDEr scores generally decrease.
To address the lack of visual information in the OFA-Speaker model, we incorporate panoramic image inputs into Models 0 and 1, resulting in Models 2 and 3. The results indicate that the CIDEr score increased by 2.97 and 4.13 points for Model 2, and by 5.24 and 5.95 points for Model 3, surpassing those of the Marky dataset. Prompt words are also found to improve language scores. To better represent spatial features between different perspectives, we design a panorama encoder based on Transformer layers and integrated it into Model 3, forming Model 4. Compared to Model 3, Model 4 achieved improvements in BLEU-4 (1.13 and 0.68 points), CIDEr (1.95 and 1.28 points), Meteor(0.8 and 0.65 points), and Rouge(0.98 and 0.64 points). These findings highlight the effectiveness of the panorama encoder in enhancing visual feature representation and instruction quality.

% \begin{table*}[tb]
% \begin{center}
%   \resizebox{0.9\textwidth}{!}{
% \begin{tabular}{ccccccccc}
% 			\toprule[1.25pt]
% 			\multirow{2}{*}{Model} & \multicolumn{4}{c}{input data}    & \multicolumn{4}{c}{val unseen}      \\
% 			\cmidrule{2-9}
% 			& crafted instruction & prompt & pano & pano encoder & Bleu4 $\uparrow$  & Meteor $\uparrow$  & Rouge $\uparrow$  & CIDEr $\uparrow$  \\
% 			\midrule
% 			Marky~\cite{DBLP:conf/cvpr/0001MOBFGJWBA22}          & \chmark    & \chmark   &     &        & 5.8   &   -   & -   & 7.5   \\
% 			\midrule
% 			0                   & \chmark    &     &     &        & 9.62  & 15.38  & 31.56 & 4.77  \\
% 			1                   & \chmark    & \chmark   &     &        & 10.97 & 15.97  & 32.39 & 4.56  \\
% 			2                   & \chmark    &     & \chmark   &         & 10.83 & 15.62  & 33.34 & 8.9   \\
% 			3                   & \chmark    & \chmark   & \chmark   &        & 12.31 & 16.47  & 34.79 & 10.51 \\
% 			4                   & \chmark    & \chmark   & \chmark   & \chmark      & \textbf{12.99} & \textbf{17.12}  & \textbf{35.43}  & \textbf{11.79} \\
% 			\bottomrule[1.25pt]
% \end{tabular}
% }
% \end{center}
%   \caption{
%   The input data did not show the results of the ablation experiment on the validation set. 
%   }
%   \label{table:r2r}
% \end{table*}

\subsubsection{Fine-Tuning Module}

Next, we evaluate the fine-tuning modules of the OFA-Speaker model. The experimental results are shown in Table~\ref{table:ft}. Model 5 utilizes the pre-trained parameters of the OFA model without fine-tuning the encoder or decoder. The low language metrics of this model indicate a significant gap between VLN instruction generation tasks and general visual-language generation tasks, highlighting the poor performance of the OFA pre-trained model when directly applied to VLN tasks.
To address this, we conducted fine-tuning experiments on the encoder and decoder separately. Model 6 fine-tuned only the encoder, while Model 7 fine-tuned only the decoder, retaining the OFA pre-trained parameters for the encoder. The results in Table~\ref{table:ft} show that Model 7 outperforms Model 6 in language metrics, demonstrating that fine-tuning the decoder alone is more effective than fine-tuning the encoder alone. This suggests that the decoder, being closer to the language generation output, plays a more critical role in improving performance.
Specifically, compared to Model 5, Model 7 achieves increases of 5.2 and 5.17 in Bleu4 scores, and 5.19 and 5.42 in CIDEr scores on the seen and unseen validation sets, respectively. Additionally, Model 4, which incorporates both fine-tuning of the decoder and other enhancements, further improves the Bleu4 scores by 5.6 and 5.24, and the CIDEr scores by 9.51 and 8.54, compared to Model 6.

\subsubsection{Decoding Method}
Finally, we investigate the decoding method used by the OFA-Speaker model, with results shown in Table \ref{table:dec}. During experiments, word repetition issues were observed in generated navigation instructions, a common problem in language generation models using greedy algorithms. To mitigate this, we adopted the SimCTG decoding method, which integrates SimCTG loss into the greedy algorithm to reduce word repetition. This adjustment led to notable improvements in language metrics. Specifically, compared to Model 8 using the greedy algorithm, Model 4 achieves increases of 0.85 and 1.27 points in BLEU scores and 1.14 and 1.91 points in CIDEr scores on the seen and unseen validation sets, respectively. Improvements are also noted in Meteor and Rouge scores.

\section{Conclusion}
This paper introduces the Generative framework for Navigation Instruction annotated with Fine-grained Cross-modal Alignment (FCA-NIG), a novel approach designed to tackle the critical challenge of data scarcity in high-quality fine-grained cross-modal alignment for VLN. By leveraging advanced foundation models such as GLIP, OFA, and CLIP, FCA-NIG automates the creation of large-scale, high-quality datasets with dual-level alignment annotations—sub-instruction sub-trajectory matching and entity-landmark correspondence. 
The proposed framework introduces a six-step pipeline: trajectory segmentation, landmark detection, crafted instruction construction, R2R-like instruction generation, entity selection, and sub-pair integration. These steps synergistically generate sub-instruction-sub-trajectory pairs enriched with entity-landmark annotations, which are aggregated into complete instruction-trajectory pairs. The resulting FCA-R2R dataset demonstrates significant improvements in navigation performance when integrated with state-of-the-art VLN models (SF, EnvDrop, RecBERT, HAMT). Specifically, the FCA-R2R dataset enhances state-awareness and action decision-making through sub-pair and entity-landmark alignment supervision (e.g., via Babywalk, AFAC, and GELA methods), achieving SR scores of 71.1\% and 68.3\% and SPL scores of 65.5\% and 63.3\% on unseen validation and test sets, respectively. 
By systematically addressing data scarcity and alignment precision, FCA-NIG advances the field of VLN toward more interpretable and scalable solutions. Future work could explore extending this framework to dynamic environments or integrating multi-turn dialog systems to further bridge the gap between simulated training and real-world deployment.

\section{Acknowledgments}
This work was supported in part by the grants from the National Natural Science Foundation of China under Grant 62332019, the National Key Research and Development Program of China (2023YFF1203900, 2023YFF1203903), sponsored by Beijing Nova Program (20240484513).
\bibliographystyle{cas-model2-names}

\bibliography{cas-refs}



\end{document}